\documentclass[runningheads]{llncs}
\usepackage[T1]{fontenc}

\title{Self-Supervised Polyp Re-Identification in Colonoscopy}
\author{Y. Intrator \and N. Aizenberg \and A. Livne \and E. Rivlin \and R. Goldenberg}

\institute{Verily Life Sciences}

\usepackage{graphicx}
\usepackage[square,sort,comma,numbers]{natbib}
\usepackage{pgfplots}
\usepackage{pgfplotstable}
\usepackage[export]{adjustbox}[2011/08/13]
\usepackage{dsfont}
\usepackage{mathtools}
\usepackage{subfig}
\usepackage{siunitx}
\usepackage{multirow}

\begin{document}

\maketitle




\begin{abstract}
Computer-aided polyp detection (CADe) is becoming a standard, integral part of any modern colonoscopy system. A typical colonoscopy CADe detects a polyp in a single frame and does not track it through the video sequence. Yet, many downstream tasks including polyp characterization (CADx), quality metrics, automatic reporting, require aggregating polyp data from multiple frames. In this work we propose a robust long term polyp tracking method based on re-identification by visual appearance. Our solution uses an attention-based self-supervised ML model, specifically designed to leverage the temporal nature of video input. We quantitatively evaluate method's performance and demonstrate its value for the CADx task.

\end{abstract}
\section{Introduction}
Optical colonoscopy is the standard of care screening procedure for the prevention and early detection of colorectal cancer (CRC). The primary goal of a screening colonoscopy is polyp detection and preventive removal. It is well known that many polyps go unnoticed during colonoscopy~\cite{van2006polyp}. To deal with this problem, computer-aided polyp detector (CADe) was introduced~\cite{livovsky2021detection, ou2021polyp, pacal2021robust, LACHTER2023} and recently became commercially available~\cite{brand2022frame}. The success of polyp detector sparkled the development of new CAD tools for colonoscopy, including polyp characterization (CADx, or optical biopsy), extraction of various quality metrics, and automatic reporting. Many of those new CAD applications require aggregation of all available data on a polyp into a single unified entity. For example, one would expect higher accuracy for CADx when it analyzes all frames where a polyp is observed. Clustering polyp detections into polyp entities is a prerequisite for computing such quality metrics as Polyp Detection Rate (PDR) and Polyps Per Colonoscopy (PPC), and for listing detected polyps in a report.

One may notice that the described task generally falls into the category of the well known multiple object tracking (MOT) problem~\cite{bytetrack, yu2022end}. While this is true, there are a few factors specific to the colonoscopy setup: (a) Due to abrupt endoscope camera movements, targets (polyps) often go out of the field of view, (b) Because of heavy imaging conditions (liquids, debris, low illumination) and non-rigid nature of the colon, targets may change their appearance significantly, (c) Many targets (polyps) are quite similar in appearance. 
Those factors limit the scope and accuracy of existing frame-by-frame spatio-temporal tracking methods, which typically yield an over-fragmented result. That is, the track is often lost, resulting in relatively short tracklets (temporal sequences of same target detections in multiple near-consecutive frames), see Supplementary Fig.~\ref{fig:tracklets}. 

A recently published method~\cite{biffi2022novel} addresses this limitation by combining spatial target proximity and visual similarity to match a polyp detected in the current frame to "active" polyp tracklets dynamically maintained by the system. The tracklets are built incrementally, by adding a single frame detection to the matched tracklet, one-by-one. However, this approach limits itself to use of close-in-time consistent detections, and cannot handle the frequent cases where polyp gets out of the field of view and long range association is required. 

In this work we propose an alternative approach that allows polyp detections grouping over an extended period of time, relaxing the spatio-temporal proximity limitation. It involves two steps: (I) a short-term multi-object tracking, which forms initial, relatively short tracklets, followed by (II) a longer-term tracklets grouping by appearance-based polyp re-identification (ReID).
As the first step can be done by any generic multiple object tracking algorithm (e.g. we use a tracking by detection method~\cite{bytetrack}), in this paper we focus on the second step.

To avoid manual data annotation, which is extremely ineffective in our case, we turn to self-supervision and adapt the widely used contrastive learning approach~\cite{simclr} to video input and object tracking scenario.

As tracklet re-identification is a sequence-to-sequence matching problem, the standard solution is comparing sequences element-wise and then aggregating the per-element comparisons, e.g. by averaging or max/min pooling~\cite{seeland2021multi} - the so-called late fusion technique. We, on the other hand, follow an early fusion approach by building a joint representation for the whole sequence. We use an advanced transformer network~\cite{attention} to leverage the attention paradigm for non-uniform weighing and "knowledge exchange" between tracklet frames.

We extensively test the proposed method on hundreds of colonoscopy videos and evaluate the contribution of method components using an ablation study. Finally, we demonstrate the effectiveness of the proposed ReID method for improving the accuracy of polyp characterization (CADx).

\vspace{1.5ex}
\noindent{To summarize, the three main contributions of the paper are:}
\vspace{-1.5ex}
\begin{itemize}
    \item An adaptation of contrastive learning to video input for the purpose of appearance based object tracking.
    \item An early fusion, joint multi-view object representation for ReID, based on transformer networks.
    \item The application of polyp ReID to boost the polyp CADx performance.
\end{itemize}

\section{Methods} \label{methods}
This work assumes the availability of an automatic polyp detector. Quite a few highly accurate polyp detectors were recently reported~\cite{livovsky2021detection, ou2021polyp, pacal2021robust}, detecting (multiple) polyps in a single frame. Our ultimate goal is to group those detections into sets corresponding to distinct polyps. 

As briefly mentioned above, the proposed approach starts with an initial grouping of polyp detections using an off-the-shelf multiple object tracking algorithm. Such a tracker is expected to track polyps through consecutive frames as long as they do not leave the camera field of view, forming disjoint, time separated polyp tracklets. In this work we use the ByteTrack~\cite{bytetrack} "tracking by detection" algorithm, but, in principle, any other tracker could be used instead. 

The resulting tracklets are typically relatively short, and there are quite a few tracklets corresponding to the same polyp. To improve the result, we propose an Appearance-based Polyp Re-Identification (ReID), which groups multiple disjoint tracklets by their visual appearance into a joint tracklet, associated with a single polyp. In what follows we describe in detail the proposed ReID component.

As stated above, the objective of ReID is to ascertain whether two time-separated, disjoint tracklets belong to the same polyp. To this end we seek a tracklet representation that allows measuring visual similarity between tracklets. The two basic alternatives are either a single representation for the whole tracklet, or a sequence of single-frame representations for each tracklet frame. 
We will consider both options below.

\subsection{Single-Frame Representation for ReID}
To generate a single frame representation we train an embedding model that maps a polyp image into a latent space, s.t. the vectors of different views of the same polyp are placed closer, and of different polyps  away from each other~\cite{hermans2017defense}. 

A straightforward approach to train such model is supervised learning, which requires forming a large collection of polyp image pairs, manually labeled as same/not same polyp~\cite{ahmed2015improved}. Such annotation turned out to be inaccurate and expensive. In addition, finding hard negative pairs is especially challenging, as images of two randomly sampled polyps are usually very dissimilar.

Hence, we turn to SimCLR~\cite{simclr}, a contrastive self-supervised learning technique, which requires no manual labeling. In SimCLR the loss is calculated over the whole batch where all input samples serve as negatives of each other and positive samples are generated via image augmentations. Combined with the temperature mechanism this allows for hard negative mining by prioritizing hard-to-distinguish pairs, resulting in a more effective loss weighting scheme.

\begin{figure}
  \centering
  \subfloat[a][]{\includegraphics[clip,width=0.173\columnwidth]{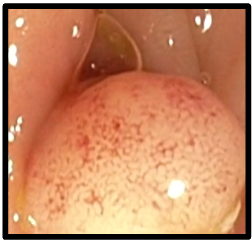} \label{fig:polyp_image} \hspace{0.2cm}}
  \subfloat[b][]{\includegraphics[clip,width=0.35\columnwidth]{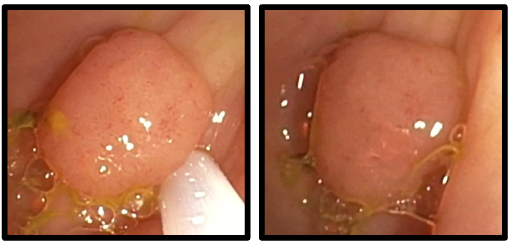} \label{fig:vid_aug} \hspace{0.2cm}}
  \subfloat[c][]{\includegraphics[clip,width=0.35\columnwidth]{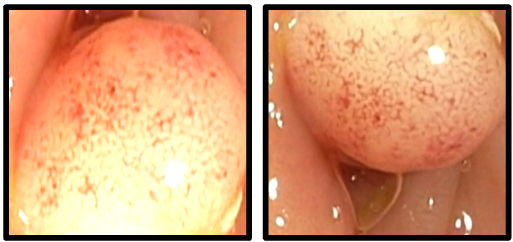} \label{fig:aug}}
  \caption{(a) A polyp image, (b) two additional views of the polyp in (a) taken from the same tracklet, (c) two typical augmentations of the polyp in (a). Images in (b) offer more realistic variations, such as different texture, tools, etc.} \label{fig:augmentations}
\end{figure}

One caveat of SimCLR is the difficulty to generate augmentations beneficial for the learning process~\cite{simclr}. Specifically for colonoscopy, the standard image augmentations do not capture the diversity of polyp appearances in different views (see Fig.~\ref{fig:augmentations}(c)).

Instead of customizing the augmentations to fit the colonoscopy setup, we leverage the temporal nature of videos, and take different polyp views from the same tracklet as positive samples (see Fig.~\ref{fig:augmentations}(b)).

Formally, a batch is formed by sampling one tracklet from $N$ different procedures to ensure the tracklets belong to different polyps. Two polyp views $i, j$ are sampled from each tracklet as positive pairs (same polyp). Let $f$ be the embedding model.
The loss function for the positive pair $(i,j)$ is defined as:

\begin{equation}
\mathcal{\ell}_{i,j} = -log \frac{exp(sim(f(i),  f(j))/\tau)}{\sum^{2N}_{k=1} \mathds{1}_{k\neq i} exp(sim(f(i), f(k))/\tau)}
\end{equation}
\noindent where $sim$ is the dot product and $\tau$ is the temperature parameter~\cite{wang2021understanding}. The final loss is computed across all positive pairs in the batch.

Tracklets represented as sequences of per-frame embeddings can be matched by computing pair-wise distances between frames, followed by an aggregation - e.g. min/max/mean distance~\cite{he2021dense, breckon2021not}. An example of similarities between frames can be seen in Supplementary Fig.~\ref{fig:heat}.

\subsection{Multi-View Tracklet Representation for ReID}
As discussed earlier, an alternative to the single frame approach, is a unified representation for the whole tracklet.
A commonly used practice is to compute single frame embedding (for each view) and fuse them~\cite{seeland2021multi, gao2018revisiting}, e.g. by averaging. The downside of those simple techniques is that they treat every frame in the same way, including bad quality, repeating, non-informative views. 
We postulate that learning a joint embedding of multiple views in an end-to-end manner will produce a better representation of the visual properties of a polyp, by allowing "knowledge exchange" between the tracklet frames.

\begin{figure}
\centering
\includegraphics[width=0.6\textwidth]{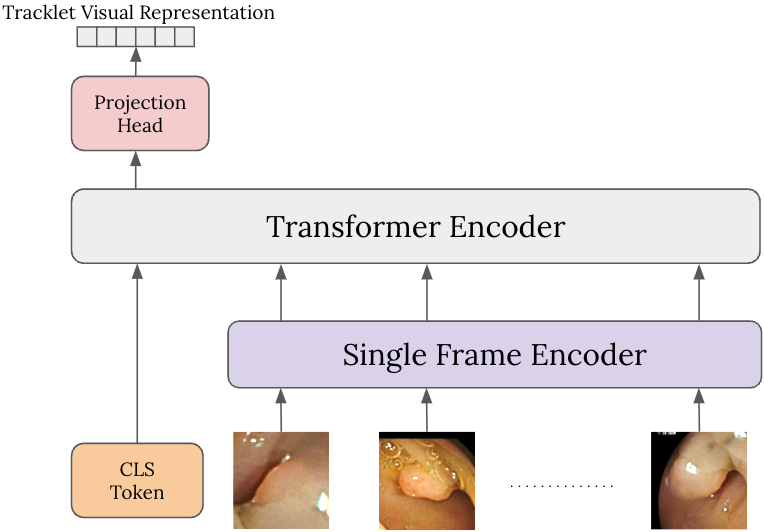}
\caption{Multi-view transformer encoder. Tracklet frames are passed through a single frame encoder to generate frame embedding. The embeddings then go through the transformer encoder, concatenated with the CLS token. Finally, the contextualized CLS token from the transformer encoder output goes through a projection head, resulting with the tracklet visual representation.}
\label{fig:multiview_transformer}
\end{figure}

To achieve this, we employ a transformer network~\cite{attention}, with the addition of BERT~\cite{bert} classification token (CLS). The attention mechanism enables both frame based intra attention and selective weighting of the frames thus providing a more comprehensive tracklet representation. The overview of the architecture is presented in Fig.~\ref{fig:multiview_transformer}. 
Training this multi-view encoder is done similarly to training a single-view encoder using SimCLR, but now, instead of pairs of frames, we deal with pairs of tracklet. 

To generate positive tracklet pairs, we cannot apply the trick used for single frames, where positive pairs are sampled within the same tracklet. Instead we generate "pseudo positive" pairs from existing tracklets. We artificially split a tracklet into 3 disjoint segments, where the middle segment is discarded, and the first and the last segments are used as a positive pair, thus providing sufficiently different appearances of the same polyp as would happen in real procedures.


\section{Experiments}
This section includes two parts. The first provides a stand-alone evaluation of the proposed ReID method. The second assesses the impact of  ReID on polyp classification accuracy.

\subsection{ReID Standalone Evaluation}

\subsubsection{Dataset}
We use \num[group-separator={,}]{22283} colonoscopy videos from 6 hospitals, split into training (\num[group-separator={,}]{21737}) and test (546) sets. For training, we automatically generated polyp tracklets using automatic polyp detection and tracking as described in Section~\ref{methods}. To reduce noise, we filtered out tracklets shorter than 1 second or having less than 15 high confidence detections, and took only the longest tracklet from every procedure. This yielded the training set of \num[group-separator={,}]{15465} tracklets (mean duration of 377 frames or 29 seconds). For evaluation, the test set polyp tracklets were manually annotated (timestamps and bounding boxes) by certified physicians. In addition, tracklet pairs from the same procedure were manually labeled as either belonging to the same polyp or not. This yielded 348 negative and 252 positive tracklet pairs.

\subsubsection{Training} 
We utilize ResNet50V2~\cite{he2016identity} as the single frame encoder, pre-trained on ImageNet~\cite{deng2009imagenet}, with an MLP head projecting the representation into a 128-dimensional embedding vector. The multi-view encoder consists of 3 transformer encoder blocks with an MLP projection head. We use LARS optimizer~\cite{lars} with the learning rate of 0.01 and $\tau=0.1$ as suggested in~\cite{simclr}. The batch size is set to 1024 for training both the single frame and the multi-view encoder. 

We first train the single frame encoder and use its weights to initialize the single frame module of the multi-view encoder. Due to memory limitations, we use 8 views per tracklet during training, resulting in $1024 * 8 = 8192$ images per training step. The model was trained for 5,000 steps using cloud v3 TPUs with 16 cores. The single frame encoder has $24M$ parameters, and the multi-view encoder adds an additional $1M$ parameters.
\subsubsection{Evaluation}
\begin{figure}
  \centering
  \subfloat[a][]{\includegraphics[clip,width=0.5\columnwidth]{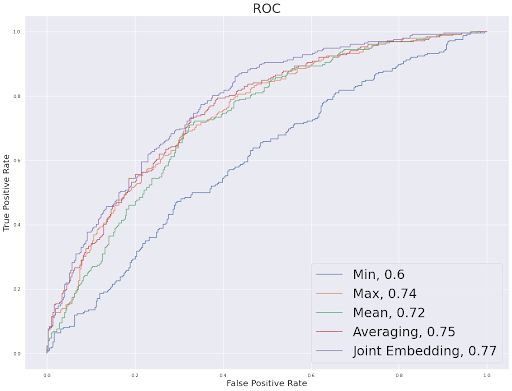} \label{fig:roc}}
  \subfloat[b][]{\includegraphics[clip,width=0.5\columnwidth]{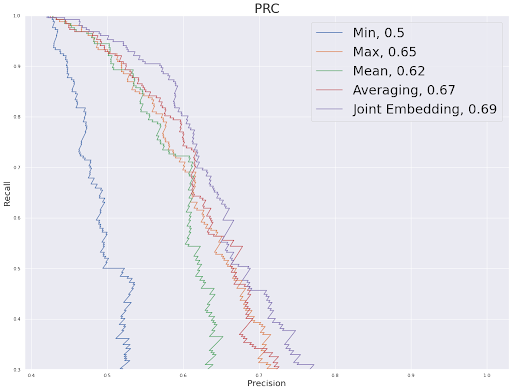} \label{prc}}
  \caption{ROC and PRC plots of various ReID techniques with AUC and AUPRC respectively. The joint embedding method consistently outperforms the other methods.} \label{fig:late_vs_early}
\end{figure}

We start by comparing various ReID techniques described in Section~\ref{methods}. Namely, we evaluate the accuracy of tracklet re-identification using: (a) single-frame representation with pairwise distances aggregation by Min / Max / Mean functions; (b) multi-view representation by frame embeddings averaging; and, finally, (c) the joint embedding multi-view model.
We evaluate the performance using AUC of the ROC and precision-recall curve (PRC) for tracklet similarity scores over the test set (see Table \ref{table:early_fusion_vs_late} and Supplementary Fig.~\ref{fig:late_vs_early}). One can see that the joint embedding multi-view model outperforms all other techniques both on ROC and PRC.

\begin{table}[t]
\centering
\begin{tabular}{|c|c|c|c|c|c|}
    \hline
    \multirow{2}{*}{} 
      & \multicolumn{3}{c|}{Single-frame} 
          & \multicolumn{2}{|c|}{Multi-view} \\            
           \hline
 & Min & Max & Mean  & Averaging &  Joint Embedding\\
\hline
AUROC & $0.60$ & $0.74$ & $0.72$ & $0.75$ & $0.77$ \\
AUPRC & $0.50$ & $0.65$ & $0.62$ & $0.67$ & $0.69$ \\
\hline
\end{tabular}
\caption{Polyp ReID accuracy for various ReID techniques.}
\label{table:early_fusion_vs_late}
\end{table}

In addition, we evaluate the effectiveness of ReID by measuring the average polyp fragmentation rate (FR), defined as the average number of tracklets  polyps are split into. 
Obviously, lower fragmentation rate means better result (with the best fragmentation of 1), but it may come at the expense of wrong tracklet matching (false positive). We measure the fragmentation rate at the operating point of 5\% false positive rate. The number of polyp fragments is determined by matching tracklets to manually annotated polyps and counting their number. Results presented in Table~\ref{table:fragmetnation_rate} demonstrate that ReID can reduce the fragmentation rate by over 50\%, compared to a tracking only solution\cite{bytetrack}.

\begin{table}[b]
\centering
\begin{tabular}{ |c|c|c|c| } 
\hline
 & FR & FR STD & Fragmented Polyps Ratio \\
\hline
Tracking & $3.3$ & $3.3$ & $0.64$ \\
Tracking and ReID & $ 1.86$ &  $1.49$ & $0.45$ \\
\hline
\end{tabular}
\caption{Fragmentation Rate (FR) statistics before and after the ReID. FR STD is the FR standard deviation. Fragmented Polyps Ratio is the percentage of polyps divided into more than one tracklet.}
\label{table:fragmetnation_rate}
\end{table}
\subsection{ReID for CADx}
In this section, we investigate the potential benefits of using polyp ReID as part of a CADx system. Polyp CADx aims to assist physicians to figure out, in real time, during the procedure, whether the detected polyp is an adenoma.

Most reported CADx systems compute a classification score for each frame, and aggregate scores from multiple frames to determine the final polyp classification. 
Grouping polyp frames into a tracklet, to be fed into the CADx, is usually done by a spatio-temporal tracker\cite{biffi2022novel}. Longer tracklets  provide more information for polyp classification.

Here, we investigate if the proposed ReID model, used to group disjoint tracklets of the same polyp, can increase the accuracy of CADx.

\subsubsection{Data} We use \num[group-separator={,}]{3290} colonoscopy videos split into train, validation, and test sets ($2666$, $296$, and $328$ videos respectively). The videos are processed by a polyp detector and tracker to form polyp tracklets. The tracklets are then manually grouped together to build a single sequence for every polyp. Each polyp is annotated by a certified gastroenterologist as either adenoma or non-adenoma.

\subsubsection{CADx}
We trained a simple image classification CNN, composed of a MobileNet\cite{sandler2018mobilenetv2} backbone, followed by an MLP layer with a sigmoid activation, to predict the non-adenoma/adenoma score in $[0,1]$, for each frame. The chosen architecture has 2.4M parameters and can run in real-time. The model was trained on Nvidia Tesla V100 GPU for 200 epochs with a learning rate of $0.001$, using Adam optimizer~\cite{kingma2014adam}. 

For evaluation, we used the model to predict the classification score for each frame and aggregated the scores using soft voting to achieve the final prediction for each tracklet. 

\subsubsection{Evaluation}
To assess the contribution of the ReID to polyp classification, we compare the CADx results on the test set, while using different grouping methods to merge multiple polyp detections into tracklets. The 3 evaluated methods are: (1) manual annotation  (2) grouping by tracking, and (3) grouping by  ReID. The manually annotated tracklets - the ground truth (GT) - are the longest sequences, containing all frames of each polyp in the test set. In grouping by tracking, we use tracklets generated by the spatio-temporal tracking algorithm~\cite{bytetrack}.
Finally, for ReID, we merge disjoint tracklets by their appearance using the ReID model. By construction, tracklets generated by methods (2) and (3) are subsets of the corresponding manually annotated GT tracklet, and are assigned its polyp classification label. A visualization of the resulting tracklets using different grouping methods is provided in Supplementary Fig.~\ref{sup-fig:data_split}. 
The number of resulting tracklets in the test set for each grouping method and polyp labels distribution are summarized in Table~\ref{table:ob_data_dist}.

\begin{table}
\centering
\begin{tabular}{ |c|c|c|c|c|c|c| } 
\hline
Grouping & Tracklets & FR & Adenoma & Adenoma FR & Non-Adenoma & Non-Adenoma FR \\
\hline
Annotation & $608$ & $1.0$ & $464$ & $1.0$ & $144$ & $1.0$ \\
Tracking & $3161$ & $5.20$ & $2537$ & $5.47$ & $624$ & $4.33$\\
ReID & $1023$ & $1.68$ & $813$ & $1.75$ & $210$ & $1.46$\\
\hline
\end{tabular}
\caption{CADx test data distribution and fragmentation rate (FR).} 
\label{table:ob_data_dist}
\end{table}

We ran the CADx model on tracklets generated by the 3 grouping methods. We compute the $F_1$ score and the AUC for the tracklet classification task. In addition, we measure the CADx sensitivity at specificity=0.9. The results are summarized in Table \ref{table:ob_resuls}. The result on the manually annotated data is the accuracy upper-bound and is brought as a reference point. One can see that the ReID based approach significantly improves the CADx accuracy compared to the tracking-based grouping.

\begin{table}
\centering
\begin{tabular}{ |c|c|c|c|c| } 
\hline
Grouping & AUC & F1 (Macro) & F1 (Micro) & Sensitivity @ Specificity=0.9 \\
\hline
Annotation &  0.95 & 0.88 & 0.91 & 0.86 \\
Tracking & 0.86  & 0.77 & 0.83 & 0.71 \\
ReID & 0.90 & 0.82 & 0.88 & 0.79 \\
\hline
\end{tabular}
\caption{Optical biopsy result per grouping method.}
\label{table:ob_resuls}
\end{table}

\section{Conclusions}
In this study we present a novel multi-view self-supervised learning method for learning informative representations of a sequence of video frames. By jointly encoding multiple views of the same object, we get more discriminative features in comparison to traditional embedding fusion techniques. This approach can be used to group disjoint tracklets generated by a spatio-temporal tracking algorithm based on their appearance, by measuring the similarity between tracklets representations. Its applicability to medical contexts is of particular relevance, as medical data annotation often requires specific expertise and may be costly and time consuming. We use this method to train a polyp re-identification model (ReID) from large unlabeled data, and show that using the ReID model as part of a CADx system enhances the performance of polyp classification. There are some limitations however in identifying polyps based on their appearance, as it may be changed drastically during the procedure (for example, during resection). In future work we may examine the use of ReID for additional medical applications, such as listing detected polyps in an automatic report, bookmarking of specific areas of the colon during the procedure, and calculation of clinical metrics such as Polyp Detection Rate and Polyps Per Colonoscopy.

\bibliographystyle{splncs04}
\bibliography{references}

\newpage

\thispagestyle{empty}		

\begin{title}
\centering
\huge
Supplementary Materials
\end{title}


\begin{figure}[!h]
  \centering
  \includegraphics[width=1\textwidth]{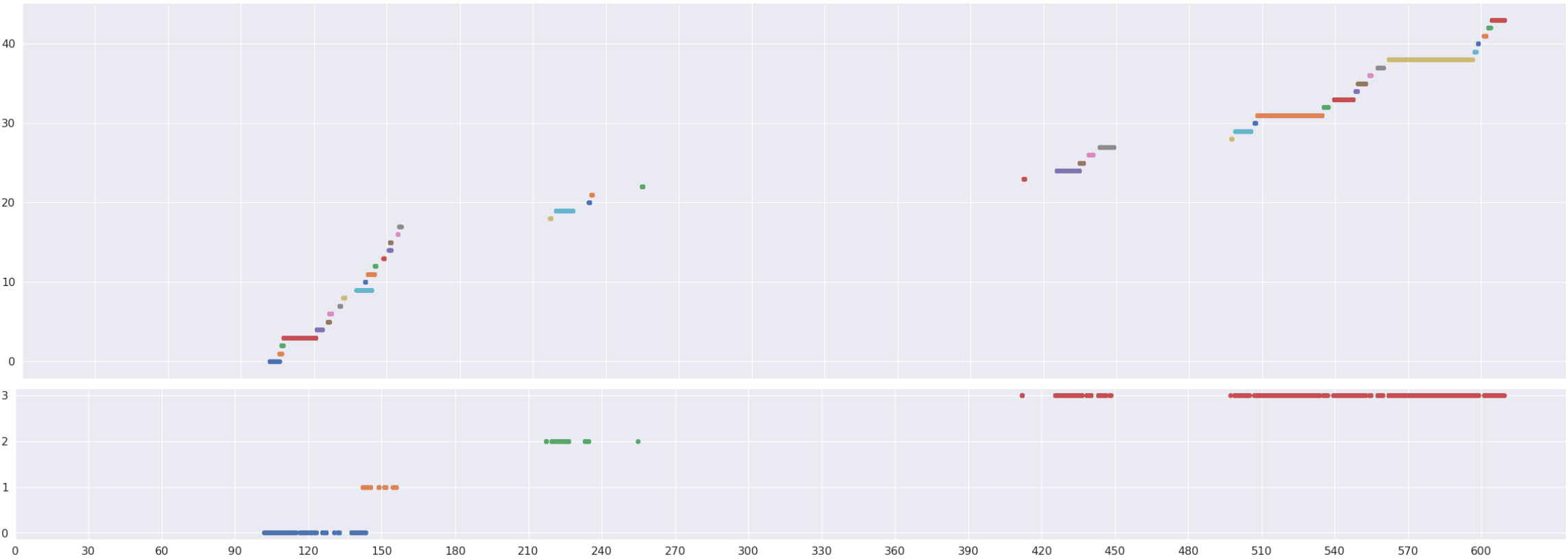} 
  \caption{Top: raw tracklets detected by tracking by detection, bottom: combined tracklets after applying ReID. X-axis is the time[s], y-axis is the unique tracklet id. Each color change represent a different tracklet. Over 40 different tracklets were found in this procedure, and only 4 remain after applying ReID.} \label{fig:tracklets}
\end{figure}

\begin{figure}[!h]
  \centering
  \includegraphics[width=0.8\textwidth]{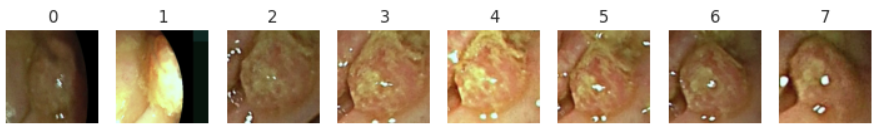} 
  \includegraphics[width=0.8\textwidth]{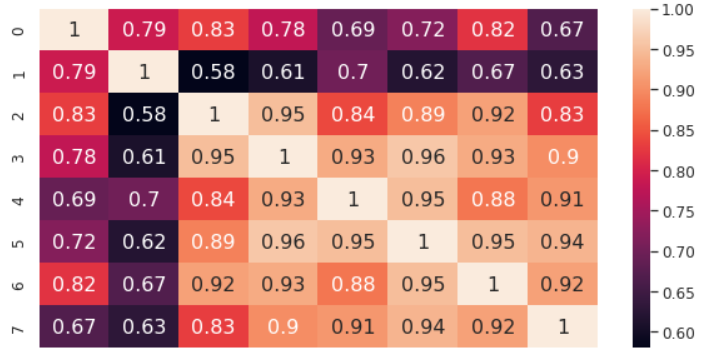} \label{fig:comparison}
  \caption{Frame-to-frame cosine similarity within a tracklet via the single frame encoder.} \label{fig:heat}
\end{figure}


\begin{figure}
\includegraphics[width=1\textwidth]{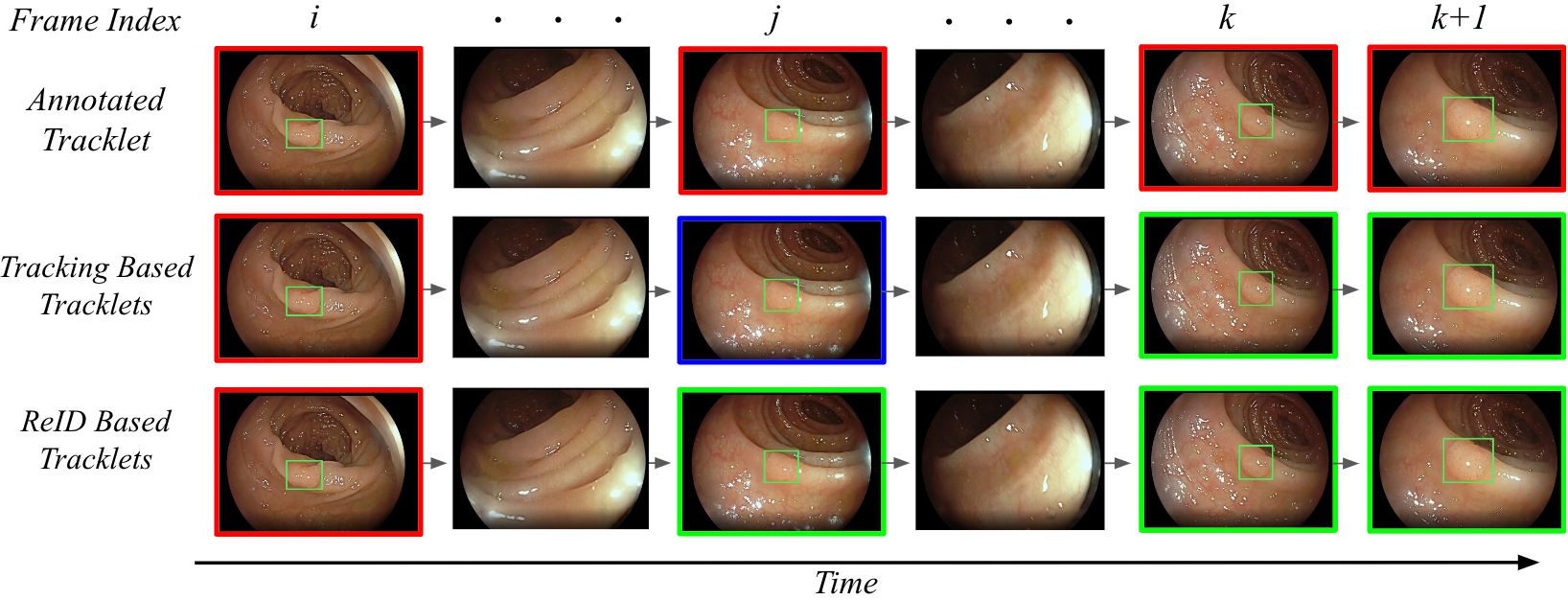}
\caption{Illustration of grouping of polyp frames into tracklets. All rows display the same frames sequence, and different borderline colors represent different tracklets. Frames without polyps appear with no colored borderline. 1st row: manual annotation - a single tracklet that includes all frames where the polyp is detected. 2nd row: the tracker groups polyp detections into 3 tracklets. 3rd row: the ReID model merged two of the 3 tracklets generated by the tracker, yielding 2 tracklets.}
\label{sup-fig:data_split}
\end{figure}

\end{document}